# Can Computers overcome Humans?

Consciousness interaction and its implications

Camilo Miguel Signorelli
Department of Experimental and Health Sciences,
University Pompeu Fabra
Barcelona, Spain
camiguel@uc.cl

Sorbonne University, Pierre et Marie Curie, Paris VI
Paris, France

*Abstract*— Can computers overcome human capabilities? This is a paradoxical and controversial question, particularly because there are many hidden assumptions. This article focuses on that issue putting on evidence some misconception related with future generations of machines and the understanding of the brain. It will be discussed to what extent computers might reach human capabilities, and how it could be possible only if the computer is a conscious machine. However, it will be shown that if the computer is conscious, an interference process due to consciousness would affect the information processing of the system. Therefore, it might be possible to make conscious machines to overcome human capabilities, which will have limitations as well as humans. In other words, trying to overcome human capabilities with computers implies the paradoxical conclusion that a computer will never overcome human capabilities at all, or if the computer does, it should not be considered as a computer anymore.

*Keywords—Artificial Intelligence; Information processing; Cognitive Computing; Type of Cognition; Super Machine; Conscious Machine; Consiousness*

I. INTRODUCTION

During many centuries, scientists and philosophers have been debating about the nature of the brain and its relation with the mind, based on the premise of an intrinsic dualism, typically called mind-body problem [1], [2]. Arguments take one form or another, however, most of them can be reduced to one kind of dualist or non-dualist view [3]. The importance of these debates acquires even more relevance when the question is stated as the possibility to build machines which would be able to reproduce some human capabilities as emotion, subjective experiences or even consciousness.

The problem is actually worst when some scientists claim a new future generation of computers, machines and/or robots which would overcome human capabilities. In view of this paper, these claims are based on misconceptions and reductionism of current most important issues. The idea, however, is not discarded here and is expressed in a different way to show its paradoxical consequences. For example, the idea of reaching and overcoming human capabilities implies the knowledge of a set of distinctive processes and characteristics which define being a human (e.g. intelligence, language, abstract thinking, create art and music, emotions and physical abilities, among others). This simple idea leads some fundamental issues. First, claims about new futurist robots do not define this set of distinctions; they do not care about the importance of what it is to be a human. Secondly, they assume a materialist view of these distinctions (i.e. these distinctions emerge from physical and reproducible interaction of matter) without explaining the most fundamental questions about the matter [4]. Thirdly, they do not explain how subjective experience or emotions could emerge from the theory of computation that they assume as a framework to build machines, which will overcome humans. In other words, these views do not explain foundations of computation that support or reject the idea of high level cognitive computers. Finally, engineering challenges of building these kinds of machines are not trivial, and futurists assume reverse engineering as the best tool to deal with this, when even some neuroscience techniques do not seem to give us any information about simple computing devices such as microprocessors [5]. Actually, if methods of neuroscience are not inferring useful information from microprocessors, it is possible to conclude that either neurons are not working as computers or all the information that we know about cells and neurons from these techniques is wrong. The first option discards reverse engineering as a feasible tool to understand the brain, and the second option discards findings in neuroscience related to mechanistic and computational interpretation. Thus, it is still necessary to focus in many intermediate and fundamental steps to declare that some computers would reach or overcome human capabilities.

This work does not expect to solve these issues; on the contrary, the aim of this paper is briefly to put in evidence misconceptions and misunderstanding of some crucial concepts. Additionally, the importance of new concepts and ideas will be approached in a preliminary and speculative way, with the intention of expanding them in further works. Following this framework in order to make clear some of the questions above, the first section will define what will be understood by human capabilities and human intelligence; the second section will confront current common view of computation, cognitive computing and information processing; the third section will discuss a basic requirement to make computers with similar human intelligence; the fourth section will show a new hypothesis of how this requirement could work; then, machines will be classified in four categories based on four types of cognitions derived from this requirement, and finally the last section will show some paradoxes, which emerge from the idea to make machines-like-brains reaching and overcoming humans.

## II. Human Capabilities

Usually, it is considered that computers, machines and/or robots will eventually reach, or even overtake human intelligence. This idea is supported by many advances in Artificial Intelligence (AI). For example, consecutive victories of DeepMind project versus the GO human champion [6], or robots that have passed some kind of Self-Consciousness test [7]. Science fiction, movies and writers also stimulate and play enough with the notion of "Singularity", the precise moment where machines exceed human capabilities [8]. In this scenario, a computer/machine is called Super Machine.

Nevertheless, how much does scientific evidence support this idea? Can computers really overcome human intelligence? What does human intelligence mean? Computers already exceed human algorithmic calculations, among many others. In fact, one option to overcome human abilities might be a cognitive system completely different to the anthropocentric science fiction view. As it will show later, this kind of computer may overcome some, but not all, human capabilities. That is why; one position could claim that it is not necessary to assume computers like brains to overcome human capabilities. It is a valid point; however, will this kind of computer overcome human brain only in a rational/algorithmic way or also an emotional one? Will this kind of computer be able to dance better than us, to create better than us, to feel better and like us? Otherwise, it will never overcome human abilities. One reason is because part of being human is to have emotional behaviour, to be able to dance, create, etc, additionally to our apparently rational behaviour. In fact, emotional behaviour might be more important for a definition of human being than rational behaviour. As it was mentioned above, the first issue emerges: what does human being mean? It is not possible thinking to overcome human abilities if it is not understood what it is to be a human and which abilities need to be overcome. For example, human intelligence may not be only associated with logical, algorithmic or rational thinking. Types of intelligence have already been suggested, which are closely related to each other such as kinaesthetic and emotional intelligences in humans [9], [10]. So far, implementing emotions or simple movements in machines is equal or more complicated than implementing rational or algorithmic intelligences [11]. Actually, current implementations of emotions in machines are based on a logical, computable and deterministic way, leaving out essential characteristics of emotions such as that emotions interfere with rational processes and optimal decisions. In fact, these implementations are based on the idea that emotions play an important role in becoming humans more efficient, rationally speaking [12], when cognitive fallacies are showing the contrary [13], [14] and experiments on neuroscience from the called default neural network, which is related with self-oriented information, are suggesting anti-correlated subsystems of information processing [15]–[17] which interfere between them. The view of computer non-like-brain does not care about this and assumes intelligence as only rational, logic and computable capability; or even worst, the problem of computer non-like-brains defenders is to think that some properties of life could be replicated without the distinctive properties of being alive.

For the purpose of this work, we will need to assume that there is certain set of "human being" properties and it is possible to decide when an animal or machine reach or not the condition to be part of this set, even when it is known that the definition of this set is one of the most controversial and debated issues.

## III. Computation and Information Processing in The Brain

One supporting fact about the idea of reaching and overcoming human capabilities with computers comes from the exponential increase of computational capacity or Moore's law [18]. This increase should impact on the development of new technologies, until reaching intelligence levels of human brain. Beyond this thought, there is the assumption that the brain works as a computer and its processing could work by analogy with computational processes. Of course, the brain is a physical entity as computers are; it partially works with electrical signals, resolves complex problems and is processing information in one way or another. Nevertheless, the way how the brain processes information is still unknown and at least, it may not be necessarily digital computation, or even more, maybe it could not be information processing in computational abstract terms at all [19]. Information processing implies processes where input are changed to become outputs; however the brain could be working in a new regime, where distinction between inputs and outputs could not exist, even causalities could be completely different to what we know until now. In this context, it should be possible to speak about another kind of processing as "replication processing", which could be self-informative to some singular physical systems like brains. It also known that brains work with complex neuromodulation [20], storages information in a sparse and unknown way [21], [22], and most distinctive yet: complex properties as subjective experiences, emotions, consciousness [23]–[25] and biased behaviour [13], [26]–[28] emerge from the brain. These emergent properties do not have relation with higher or lower computational capability. For example, the cerebellum has more neurons than any other part of the brain, but it does not play any important role in conscious perception [29].

Related with this notion, a common assumption in cognitive science is to consider processing of information as synonym of computation; however it is necessary to differentiate both concepts. For instance, if information is considered as the content of a message, this content would need a physical system to be propagated and stored. Thus, information may be understood or at least associated with a physical entity [30]. According to general view, information processing can be any physical process which transforms an

This work was partially supported by Comisión Nacional de Investigación Ciencia y Tecnología (CONICYT).

input into an output. Information processing can also be defined in terms of causality between inputs and outputs. Additionally, computation could be understood as syntactic and symbolic manipulation of information [2]. In this sense, computation is an algorithmic and deterministic type of information processing. Although it is possible to appeal to a non deterministic computation, in general, this non deterministic computation can be reduced to deterministic types of simple computation at level of Turing test. The problem is that brains are not just doing computation, they are also able to give interpretations and meaning to their own high level information processing. Arguments in favour of this idea are stated from philosophical view in [2] and psychological/biological view in [25].

One interesting case of computation is artificial neural networks, which could be interpreted as semi-deterministic information processing systems. Artificial neural networks evolve in a non deterministic way thanks to self-learning and trained from some given rules, which are not always explicitly programmed. Since artificial neural network as for example Hopfield networks [31] are inspired by biological principles [32], [33], which are in turn inspired by biological observations [34], one option to introduce semantic and meaning to artificial networks could be implementation of interactions between subsystems as observers of each other in a context of artificial neural networks. It will be discussed in next sections. Through this way, intelligence would not be only associated with deterministic logical computation but with the interaction between deterministic, semi-deterministic, non-deterministic, and perhaps quantum computation, or even new frameworks of processing of information.

While some computer and cognitive scientist might not agree with this interpretation of information and computation, it is still admissible to have processing of information without computation and intelligence without a deterministic way of processing of information. Actually the brain apparently does it. In fact, the most important features of the brain are the result of unpredictable, nonlinear interactions among billions of cells [35], [36]. Science does not know which are the real decodes and encodes of the brain; does not know as cognitive abilities emerge from physical brains, and even more complicated, it is not certain that we have a deterministic way to explain how this emergency works.

At this point, the usual idea of digital computation in cognitive science and neuroscience should change in favour of a perspective of computation and information processing by analogy with physical systems where inputs, rules and outputs can be interpreted in a physical and global way.

The brain should not be thought as a digital computer neither in the "software" [1], [2] nor in the "hardware" [19], [37], [38]. One reason is because this analogy obscures the complex physical properties of the brain. On the one hand, neuroscience and cognitive science use indiscriminately concepts as information, computation and processing of information without understanding of their physical counterpart, sometimes based on the assumption of not hardware dependency of these concepts, other times due to assumptions of how the brain encodes and decodes information. The most common assumption is to think that activation or spikes in neurons is the only informative state. While other cells as astrocytes [39], and non classical integration as neuromodulatory substances [20], back-propagation [40], among others [38] are ignored. In addition, inactivation and deactivation states could also be encoding valuable information about dynamical brain states at macro and micro scale. Neurons never are in static state and their membranes are presenting fluctuations that could still be informative. The distinctive physical brain properties and their dynamical interactions are apparently more important than computational views were thinking, which implies that hardware can not be ignored at all. According to this point, the analogy between a drum and the brain could be more informative than the analogy brain-computer. Drums can respond with different and complex vibration states when they are stimulated, and they can be also understood on computational terms: input (hits), rules (physical laws, physical constrains such as material, tension, etc), and outputs (vibration, sounds, normal modes). Indeed, the brain has many more similarities with a dynamical system as a drum than with digital computers, which are based on discrete states. Drums, as well as brains, are dynamical systems with emergent and sub-emergent properties, drums have different modes of vibration, superposition, physical memory, sparse storage of this memory, among others features. In abstract terms, drums are also computing and processing information, but this information processing is a dynamical reaction from external stimuli more than a formal calculation process (computation as defined above).

On another hand, computer science is losing valuable information on the attempt of replicating brain capabilities. One example could be alpha, gamma or oscillations of brains in general [41], synchrony [42], [43], harmonic waves [44], among other processes which are not seriously considered in artificial intelligence, neither using artificial neural networks. This characteristic should be understood and incorporated in order to implement desirable social behaviour in new generations of computers, machines and robots. Considering that some of these behaviours are intrinsically related to biological organisms, perhaps these behaviours are not reproducible without some intrinsic constituents of information processing of biological organisms [45], [46] as for example oscillations.

Finally, abstractions and general concepts are really useful in theoretical terms; however concepts as computation, information and information processing in the brain do not have evident interpretation. Realizing that these concepts should not be used as an analogy with computers is the only way to lead us to the correct direction: Focusing on differences between brains and computers, and trying to fill the gaps without assumptions. Maybe, for many computer scientists, these comments are trivial, but what computation means for computer science is not the same than for biological science, leading misunderstandings and misconceptions, while also the knowledge that computer sciences have about codification in the brain is very limited, leading to erroneous assumptions.

To sum up, these two section have identified some usual presumptions: i) The assumption of a set of distinctive properties defining human being, ii) intelligence related only to logical and rational thinking, iii) brains working by analogy with hardware-independent computers, iv) computation as synonym of information processing, and v) brain information only encoded in activation states of neurons. When differences between concepts appear, it becomes necessary to clarify some of them. For example, a better understanding and definition of information processing in the context of human intelligence, where computation will be a kind of information processing between many other types, including the characteristic one to biological organisms [46]. Probably, new concepts and foundations of information will also be needed, especially to understand the real encode and decode of brain cells, as a crucial theoretical starting point. These foundations should be intrinsically related to minimal constitutive parts of physical theories. Thus, computer-brain analogy is not useful anymore, at least in the current sense. Nevertheless, it could still be possible to replicate some brains abilities thanks to new formulations of information processing and theoretical frameworks.

## IV. Consciousness as Requirement

Intelligence should also be considered as a whole. Intelligence is often understood as the ability to solve problems in an efficient way, thanks to other mechanisms as learning and memory. This means, the maximization of the positive results in our solution while minimizing the negative impacts, for instance, waste of time. To do that, other processes as learning and memory are also needed and associated with the definition of intelligence. In a general sense, learning may be understood as the process to gain new knowledge or improve some behavior, while memory is the storage of this knowledge. To solve problems efficiently, it is necessary to access a certain memory that was acquired thanks to a specific learning that will be modifying again the memory of the system. The more intelligent is the system, the more it learns. However, in this framework it is forgotten that emotions as subjective experiences, and cognition are deeply related to human intelligence [36]. They play a crucial role in learning, in consolidation of memories, retrieved memory and in human cognition in general [25].

Furthermore, one requirement for emotional and logical/rational intelligences seem to be what is called subjective experience [47] or in a more complex order: Consciousness. Humans first need to be conscious to take some complex rational decisions, to plan, and to have the intention to do something [48], [49]. For example, vegetative patients and minimal conscious patients do not present signals neither planning nor having intentions to do minimal tasks [50], even when they could present minimal signs of consciousness [51]. Planning and intentions apparently emerge when minimal signs of consciousness exceed a threshold. In fact, these minimal signs can be interpreted as predictors of recovering in minimal conscious patients [52], [53]. Other works are re-defining the idea of subjective experience until its minimal constitutive part and argue the existence of minimal subjective experience even in insects [47]. It could mean that complex decisions, planning and having intentions are different from consciousness, although they are closely related: Subjective and conscious perceptions are apparently previous to rational intelligences, planning and even to efficient behaviors. Experiments in psychology of judgment and behavioral economics have also shown as subjects tend to perform some tasks in a biased way even when they have been trained, suggesting that logical and rational intelligences appear only after more elaborated information processing [13], [14]. It is clear that how biology implements high intelligences is completely different from how computer science implements it [11].

The need to incorporate subjective experience and eventually consciousness implies a complex problem which involves many different processes as awareness, emotions, subjectivity, intentionality, and attention, among others. Consciousness should be formed from all of these processes like a differentiated and unified whole, but it is not any of them. For example, it could be necessary to be aware to have emotions and subjective experiences, or maybe vice versa, and we will need them to show intentionality, attention and high level cognitive abilities. It is also necessary to insist and distinguish that these are different processes, for example awareness and attention; while it is important understanding all of them as constituent parts of what we define as consciousness.

Together, these ideas could imply that to reproduce high level of human intelligence in a biological way, it is necessary to introduce subjective and conscious behaviour in machines at early stages. As subjective experiences and conscious behaviour are closely related with emotions, and emotions are related with learning and intelligence in humans, the only way to overcome human brains would be making conscious machines which would be able to reproduce emotional human intelligence, in addition to logical intelligence. In order to implement high-level-computers, that is to say computers-like-brain, it will be necessary to focus on conscious human capabilities, and how it is impacting the information processing of the system.

## V. Consciousness Interaction Hypotheses

It is common to think that computer reaching human capabilities is only question of time. From the comments above, the question is not how much time it is necessary to reach human abilities with machines; the question is how to implement consciousness in machines, when it is not already known what consciousness is. The problem is also that consciousness can be understood in many different ways and actually, there is not any correct and complete definition about consciousness.

Instead, it is possible to identify some characteristics of consciousness, awareness and conscious perception. First, it is not matter of capacity of computation. The brain should not be considered as a computer neither doing any computation like a computer. Although, if someone would like to insist, the brain capacity can be roughly estimated in 20 petaFLOPS, assuming 100 billions of brain cells, 200 firings per second, 1000 connections each cell (see other approximations [54] ), whereas independently of any approximation, 80% of these brain cells (hence its computational capacity) are in the cerebellum, which does not play any important role in conscious perception [29].

Additionally, the most powerful computer has 93 petaFLOPS (Sunway TaihuLight [55], [56]) and it is really unlikely that someone ensures that this computer is aware because of its bigger computational capacity. Secondly, evidence has shown that conscious perception needs between 200 to 400 ms, and some experiments also suggest a discrete mechanism instead of a continuous perception mechanism [57]–[60]. For example, evidence for discrete mechanism of perception comes from psychophysical experiments where two different stimuli are presented with a short time window between each other. In these experiments, subjects perceived both stimuli as occurring simultaneously, suggesting a discrete temporal window of perception integration [57], [59]. Other experiments, where subjects are exposed to masked stimuli (words or pictures which are masked by previous stimuli), have showed that conscious perception (i.e. subjects report seeing the stimulus) is associated with a positive peak in Event-related potentials (ERPs) which appear 300-500 ms after the stimulus presentation [58], [59]. This response is called P3b and has also been related with attention and memory processes. On the contrary, the processing and integration of information at low level tasks only need 40 ms. In other words, when we perceive consciously, any processing of information is temporally decreasing between 500% up 1000%. Thirdly, it is possible to observe an apparent "interference" between different types of information processed in human conscious behaviour. For instance, rational calculations (e.g. resolve a mathematical problem) interfere with kinesthetic performance [61], supporting the idea to treat intelligence as a whole system formed by parts of processes. To illustrate, resolving a mathematical equation while biking or dancing at the same time could be practically impossible. This observation suggests that conscious perception would be imposing a balance between different processes, and hence a balance of different intelligences. Computational interpretation of this observation will try to explain the interference between different kinds of information as a competition for computational capacity. However, as it is stated above, computational capacity apparently is not playing any crucial role in perception. This analogy also assumes processing of information in a digital way, which could be not the best approach to understand the brain. Fourthly, some results from behavioral economics and decision making have shown that cognitive biases are not according with classical probability frameworks [62]. It means that it is not always possible to treat emergent brain properties in a classical and efficient probabilistic way.

If consciousness is not a matter of computation capacity whereas temporal efficiency decreases in its presence, it could be due to its architecture. Many theories have tried to explain how consciousness emerges from the brain [23], [24]. However, these theories are incomplete although they might be partially correct. The incompleteness is because most of these theories are descriptions of the phenomenon, instead of explanatory theories of the phenomenon (e.g. Classical Mechanics or Theory of evolution, although an explanatory and/or complete theory does not ensure that it is correct). Descriptive theories focus on how the phenomenon works, use descriptions without causal mechanisms even when they claim it, and without deductive general principles, i.e. they often start from the object of study to deduce specific/particular principles rather than deducing general principles and in consequence explaining the object of study. Furthermore, incomplete theories do not answer one of these fundamental questions: What is "the object of study"? How does it work? Why? Most commonly, they do not explain "why" something works as it works. In other words, these theories may partially explain how consciousness emerges, but they do not explain and do not solve the entire problem. Finally, these approaches try to explain awareness or conscious perception in a way that is not clearly replicable or implementable in any sense, neither with biological elements. Some theories also use the implicit idea of computability to explain, for example, conscious contents as the access to certain space of integration; and competition for space of computation in this space, to explain how some processes lose processing capacity when we are conscious.

Another complementary alternative is to understand consciousness as intrinsic property due to the particular form of information processing in the brain. Here, consciousness will be interpreted in this way, as the "interference", or "interaction" of different neural networks dynamics, trying to integrate information to resolve each particular network problem. More specifically, the brain could be divided into different "principal layers" (topologically speaking) which are also formed by different levels of layers, each principal layer as one kind of neural network interconnected at different levels with other networks. Each principal layer can process information independently of other principal layers; however when they are activated at the same time to resolve independent problems, the interaction generates a kind of interference on each intrinsic process. From this interaction and interference would emerge consciousness as a whole. I will call it: Consciousness interaction hypothesis or consciousness interference hypothesis. Consciousness would be a process of processes which mainly interferes with neural integration. These processes are indivisible part of consciousness, and their interference is the consciousness. This hypothesis might allow us to explain why the brain is not always an efficient machine, why decisions are not always optimal, why it is possible to observe an apparent loss of processing capacity between different types of information processing in human conscious behaviour, and more interesting, it allows us to implement a mechanism on other machines than biological machines. Although these ideas still do not answer the "why" question of a complete theory of consciousness, they are part of a global framework related with codification, processing of information and category theories which will intent to respond that question and will be developed in further works. Some important differences of this framework with previous approaches are: (1) awareness would emerge from breaking neural integration, synchrony and symmetry of the system; (2) conscious perception would correspond to dynamics operations between networks more than containers formed by networks where to put contents. In this sense, consciousness is a distributed phenomenon by essence; (3) consciousness interaction hypothesis could be an implementable mechanism for artificial intelligence.

Finally, one crucial observation emerges from this discussion. Consciousness interaction hypothesis requires a balance between different processes involved in its emergence. That is why extraordinary capacities in some processes are compensated with normal or sub-normal capacities in other processes of information when we are conscious.

## VI. TYPES OF COGNITION AND TYPES OF MACHINES

Consciousness interaction is a different framework, therefore it is needed to re-interpret some definitions from previous theories about consciousness [24]. **Conscious states** as different levels of awareness (vegetative, sleep, anesthesia, altered states, aware) would correspond to different degrees of interaction or interference between different networks. In consciousness interaction hypothesis, consciousness is not a particular state neither has possible states, at difference of common definitions. Consciousness should be interpreted as an operation/process itself. **Contents of consciousness** as elements or information in the external or internal world which at times are part of our conscious perception, would correspond to superposition of different oscillation related with "intersection points" of interference between networks or the network points (nodes) which are influenced/affected by this interference/interaction (probably in a scattered/sparse way). Finally, **conscious processing** is normally defined as the operations applied to these contents. In consciousness interaction framework, it would correspond to constants "loops" of interference/interaction on this "intersection points" and its dynamic evolution.

With similar definitions (without this particular interference interpretation) and their relations, Shea and Frith have identified four categories of cognition [61] depending if contents and cognitive processes are conscious or not: Type 0 cognition corresponds to cognitive processes which are not conscious neither in contents nor operations applied to these contents. Type 1 cognition is a cognitive process where contents are consciously perceived, however operations on these contents are not controlled. Type 2 cognition would correspond to contents and operation on these contents consciously perceived and controlled. Finally, what I have called Type ∞ cognition (to see [63]) can be understood as cognition without contents consciously perceived, but operations on these contents are consciously controlled. Based on these definitions, in Table 1, it is also possible to relate these categories with four categories of machines and their information processing capabilities [63]: (1) The **Machine-Machine Type 0 Cognition**: non-conscious contents and non-conscious (non-controlled) processing. Optimal choice is computation-light but learning-heavy (e.g. Motor Control). Examples are robots that we are making today with a high learning curve. (2) **Conscious-Machine Type 1 Cognition**: conscious contents but automatic and non-controlled processing (e.g. Fallacy questions). The system accesses to a wider range of information thanks to first levels of interference/interaction between networks (Holistic information), however some optimal or specific algorithmic calculations may become intractable. (3) **Super Machine Type 2 Cognition**: contents are conscious and the cognitive process is deliberate and controlled thanks to a recursive and sustained interference/interaction at certain intersection points from different networks (e.g. Reasoning). From a computational view, it is considered computation-heavy, learning-light and interferes with Type 0. (4) **Subjective-Machine Type ∞ Cognition**: non-conscious content but conscious process (no analogy with humans because this condition has not been observed). My hypothesis about this type of machines is related to Supra reasoning information emerged from organization of intelligent parts of this supra system (e.g. Internet).

Some previous works have been also tried to generalize and characterize some features of consciousness that could be related with types of machines and/or artificial systems [64]. For example in [65], even thought this article still keeps a computational view of consciousness and social interactions, they conclude that consciousness is not only related with computational capacity and put emphasis in social interactions (which can also be related with emotions) as trigger of consciousness. Another example is [66], where some categories defined can be related with some types of machine mentioned above. Nevertheless some crucial differences with these articles are: 1) Here, types of machines directly emerge from previous theoretical and experimental definitions of types of cognition. In this context, types of machines are general categories related with definition of cognition and its relation with consciousness. 2) In this article is not assumed any especial optimization processes to reach consciousness, actually quite contrary, interference processes as not-optimal processes and some still missing properties of soft materials/brains would be related with its emergence.

Due to these not-optimal processes, each type of machines has limitations [63]. For instance, conscious machine type 1 cognition does not have strong algorithmic calculation capabilities or rational/logical intelligence, because accuracy is lost in favour of consciousness as fast access to holistic information. Subjective machines type ∞ cognition probably will not be able to interact physically with us, and even less dance like us or feel like us, however it is the most likely scenario where machines and computers will be able to overtake some humans capabilities and keeping current hardware in a non anthropomorphic way. For this machine, subjective experience could be something completely different to what it means for humans. In other words, Subjective Machines are free of human criteria of subjectivity. Eventually, Super Machine is the only chance for AI to overcome human abilities as such. This machine would have subjective experiences like humans, at the same time that it would have the option to control the accuracy of its own logic/rational process; however, it is also vulnerable to what subjective experiences imply: the impact of emotions in its performance and biased behavior as humans.

|  | Non conscious Processing | Conscious Processing |
|---|---|---|
| Non conscious contents | Machine-Machine | Subjective Machine |
| Conscious contents | Conscious Machine | Super Machine |

**Table 1.** Types of machines according different types of cognition, contents and information processing exposed above.

## VII. IMPLICATIONS FOR ARTIFICIAL INTELLIGENCE

If we can reach the gap to make conscious machine type 1 or 2 cognition, these machines will lose the meaningful characteristics of being a computer, that is to say: to resolve problems with accuracy, speed and obedience. Any conscious machine is not a useful machine anymore; unless they wanted to collaborate with us. It means the machine can do whatever it wants; it has the power to do it and the intention to do it. It could be considered a biological new species, more than a machine or only computer. More important: based on our previous sections and empirical evidence from psychology and neuroscience [36], [63], it is not possible to expect an algorithm to control the process of emergency of consciousness in this kind of machines, and in consequence, we would not be able to control them.

With this in mind, some paradoxes appear. The first paradox is that the only way to overcome human capabilities with computers is making machines which are not computer anymore. The second paradox is that when we make conscious machines type 1 and/or type 2 cognition, a process of interference, due to consciousness, will affect the global processing of information, allowing extraordinary rational or emotional capabilities, but never both extraordinary capabilities at the same time or even in the same individual. In fact, if the machine is a computer-like-brain, this system will require a human-like-intelligence that apparently requires a balance between different intelligences, as stated above. That means that a machine type 1 or type 2 cognition would never overcome human abilities at all, or if it does, it will have some limitations like humans. The last paradox, if humans are able to build conscious machine that overcomes human capabilities: Is the machine more intelligent than humans or are humans still more intelligent because we could build it?

## VIII. CONCLUSIONS

These comments seek to motivate discussion. The first objective was to show typical assumptions and misconceptions when we speak about AI and brains. Perhaps, in sight of some readers, this article is also based on misunderstandings, which would be another evidence of the imperative need for close interaction between biological sciences, such as neuroscience, and computational sciences. The second objective was tried to overcome this assumptions and expose a hypothetical framework to allow conscious machines. However, from this idea emerge paradoxical conclusions of what a conscious machine is and what it implies.

The hypotheses stated above are part of "prove of concept" to be commented and reformulated. They are part of a work in progress. Thanks to category theory and others theoretical frameworks, it is expected to develop these ideas on consciousness interaction hypothesis more deeply, and relate them with other theories on consciousness, its differences and similarities. In this respect, it is reasonable to consider that a new focus that integrates different theories is needed. This article is just the starting point of a global framework on foundation of computation, which expects to understand and to connect physical properties of the brain with its emergent properties in a replicable and implementable way to AI.

In conclusion, one suggestion of this paper is to interpret the idea of information processing carefully, perhaps in a new way and in opposition to the usual computational meaning of this term, specifically in biological science. Discussions about it and other future concepts instead of information processing into the brain should be expanded in further works. Additionally, although this work explicitly denies the analogy brain-digital-computer, it is still admissible a machine-like-brain, where consciousness interaction could be an alternative to implement desirable high intelligences in machines and robots. Even if this alternative is neither deterministic nor controlled and presents many ethical questions, it is one alternative that might allow us to implement a mechanism for conscious machine, at least theoretically. If this hypothesis is correct and it is possible to reach the gap of its implementation, any machine with consciousness based on brain dynamics may have high intelligences properties. However, some kind of intelligence would be more developed than others, because by definition, its information processing would also be similar to brains which have these restrictions. Finally, these machines would paradoxically be autonomous in the most human sense of this concept.

## *Acknowledgment*